\documentclass{article}

\usepackage{arxiv}

\usepackage[utf8]{inputenc} 
\usepackage[T1]{fontenc}    
\usepackage{hyperref}       
\usepackage{url}            
\usepackage{booktabs}       
\usepackage{amsfonts}       
\usepackage{nicefrac}       
\usepackage{microtype}      
\usepackage{lipsum}
\usepackage{graphicx}
\graphicspath{ {./images/} }

\title{Social Practices for Social Driven Conversations in Serious Games}

\author{
Agnese Augello \\
ICAR - National Research Council of Italy \\
Viale delle Scienze - Edificio 11 \\
90128 Palermo, Italy \\
\texttt{augello@pa.icar.cnr.it} \\
\AND Manuel Gentile \\ 
ITD - National Research Council of Italy \\
Via Ugo La Malfa 153\\
90146 Palermo, Italy \\
\texttt{manuel.gentile@itd.cnr.it}\\
\AND Frank Dignum \\
Utrecht University, The Netherlands\\
Princetonplein 5, De Uithof\\
3584 CC Utrecht \\
\texttt{F.P.M.Dignum@uu.nl}\\
}

\begin{document}
\maketitle

\begin{abstract}
This paper describes the model of social practice as a theoretical framework to manage conversation with the specific goal of training physicians in communicative skills. To this aim, the domain reasoner that manages the conversation in the Communicate! \cite{jeuring}  serious game is taken as a basis. Because the choice of a specific Social Practice to follow in a situation is non-trivial we use a probabilistic model for the selection of social practices as a step toward the implementation of an agent architecture compliant with the social practice model.

\end{abstract}
\section{Introduction}

Effective communicative skills are important in different fields, and are the basis of every social relation. Serious games can be useful in this context, as a valid approach to train people to properly carry out conversations by means of simulation of dialogues with virtual  characters. See e.g. \cite{traum2013}\cite{sabarish}\cite{dargis}.

The  use of virtual agents  for this purpose is particularly useful because it allows us to also bring social elements of interactions into the simulations  \cite{traum2013}. They can be used to provide learners with a continuous  feedback, showing the  effects of their conversational choices on the emotions and behavioural changes of the interlocutors \cite{jeuring} \cite{traum2005} .  
 
An area where conversations play an important role in everyday practice are the medical consultations.
Training physicians to communicate better has several positive effects on patients' well-being. For example a good communication enhances patients' treatment adherence \cite{Dimatteo2010}, leads to greater patient disclosure of sensitive psychosocial information and foster a reduction in patients' emotional distress \cite{roter1995}.
It is generally agreed that better doctor-patient interaction leads to better diagnosises, which in turn can lead to huge (financial and human) cost savings in hospital tests and treatments.

Several Serious Games have been designed in this context. E.g. \cite{jeuring} \cite{traum2015}. 
Communicate!\cite{jeuring} is a serious game used by medical students to learn the best communication strategies to establish a valid and trusted relationship with the patient by means of consultations with virtual characters. SimSensei Kiosk \cite{traum2015} , a dialogue system that conducts interviews related to psychological distress conditions, shows several benefits of using virtual agents from the perspective of the patient. In particular the patient, interacting with a virtual agent shows more openness to the conversation, not feeling judged by another human interlocutor.

Most agents that are used in these applications are based on scripts and reactive rules to respond to users. Unfortunately this leads to quite predictable and simple dialogs. Thus the challenge is to design agents that are more pro-active and are able to autonomously take their decisions according to their expectations and the evolution of the game. This would be more in line with Clark \cite{Clark} who claims that a dialogue is a joint activity that must consider both individual and social processes.

The social context has a key role in the deliberative process of agents and is particularly important in a conversational context, where the deliberative process regards the choice of the most proper utterance.  
Conversation has a social effect because it contributes to changes in  people (beliefs, attitudes, etc.),  social relations, and the material world \cite{fairclough}.
In the other hand the social structure, the social practice and the social agents involved in the social interaction determine the actual conversation utterances that are used \cite{fairclough}.
As a matter of fact different communication strategies can be used according to the specific social context. 

It is important therefore to formalize how the social context influences the deliberation process of agents. Existing techniques  often do not fully model this type of situational deliberation  of agents.  The social context is often dealt with as a set of norms that add complexity to the cognitive model of the agent, complicate the  deliberative process and  that restrict  agents application to the particular application domain  for which they have been designed \cite{dignum}.

In order to address this issue we intend to use the concept of social practice as modeled in \cite{dignum}.  Social practices refer to everyday practices and the way these are typically and habitually performed in a society.  In \cite{dignum} social practice are used  to create agents that are able to sense the physical and social aspects of the current situation, but act not in a simple reactive way, but pro-actively choosing a plan of actions suitable to reach their both social and functional goals. 

In this paper, the model of social practice as a theoretical framework to manage conversation with the specific goal of training physicians in communicative skills is analyzed. To this aim, the domain reasoner that manages the conversation in the Communicate! \cite{jeuring}  serious game is taken as a basis. We see how scenarios build with Communicate! can be replicated using Social Practices in order to subsequently show how the Social Practice model enhance the scenarios in a flexible and natural way.
Because the choice of a specific Social Practice to follow in a situation is non-trivial we use a probabilistic model for the selection of social practices as a step toward the implementation of an agent architecture compliant with the social practice model \cite{dignum}.

%

\section{Training communication skills for medics}
\label{sec:training}
Let us consider the following scenario. A doctor is at the desk of his office in the hospital where he works. Someone knocks at the door. Because it is his consultation hour he expects a patient. Thus the doctor invites the person to come in. 
He habitually sees a lot of patients, following standard conversational protocols aimed to  obtain useful information from the patient such as the existence of particular pathologies and  the symptoms of the disease.

When the doctor receives a patient he activates a plan of actions related to the conversational protocol that should lead him to understand the patient's problems and to give the patient the right therapeutic treatment. 

When the doctor does not know the patient who has just entered the room, he will first introduce himself properly and ask some general background questions to establish a trusted relationship with the patient. 

Once he starts the conversation he will have some expectations about the type of responses the patient will give. If the expectations are not met he might change his ideas about the patient and reconsider his next moves in the conversation. E.g. if a patient seems not to acknowledge a symptom or fact, it might be the patient is worried or distracted and needs first to be put at rest.
However, if	 an emergency occurs the doctor should interrupt the consultation and start a totally different practice (belonging to the emergency). If he did not handle this type of emergency very often he will have to deliberate to a greater extent (based on his medical knowledge and other cases) to determine the right plan of action.

The scenario described above shows the complexity of autonomous agents for this kind of games. 

In order to have a concrete reference model for the medical context, we start from the work done in the Communicate! project. First the scenario model defined in the Communicate! serious game is briefly introduced.
Then, the  ``anamnesis" scenario is analized in order to understand how 
scenarios build with Communicate! can be replicated using Social Practices in order to manage the conversation in a flexible and realistic way. 

\section{Communicate! dialogue management}

The scenario model defined in the Communicate! project allows the designer to specify:
\begin{itemize}
\item a set of parameters, that represent the dialogue state;
\item one or more conversation trees, in which the nodes are the player's statements and the computer's statements;
\item a sequence of interleaves, that represent a sort of dialogue phase in which one or more conversation trees could take place;
\end{itemize}
The Communicate! serious game is based on the structural dialogue state approach, in which the interleaves, the trees and the statements represent the dialogue grammar.  
The dialogue state is defined by the scenario parameters' scores, by the lasts computer's statements and by a set of emotions' scores. The basic set of emotions managed in Communicate! are : happiness, anger, surprise, contempt, disgust, fear and sadness.

According to the current dialogue state, a domain reasoner selects the possible moves (the player statements) among which  the player could choose the move to carry forward the dialogue.  In fact, the player statements represent the dialogue moves that update the elements of the dialogue state (the parameters and the emotions of the virtual character). 
\subsection{The ``anamnesis" scenario}
The situation described in section \ref{sec:training} is managed by a specific scenario defined in the framework of the Communicate! project \cite{jeuring}, the ``anamnesis" scenario. 
In this scenario, the user plays the doctor role and holds a consultation with a virtual patient. 

The dialogue is ``started" by the player who assumes the role of the doctor. 
The player has to choose among a number of possible expressions. Each of these expressions leads to the activation of a conversation tree.

Let us suppose the player, among the different possibilities, chooses the following sentence to proceed the scenario:
``I see you are a patient of Dr. Aarts"
This player statement, produces a ``surprise" effect in the  agent that realizes that he will have an interview with an unknown doctor.

In terms of social practice, this element of ``surprise" is the result of a violation of an expectation of the social practices activated by the patient. Infact, as shown to the player at the begginig of the game, the patient expects to have an interview with his own doctor, that is the patient agent has activated the social practice \textit{consulting my doctor}. 
This lead the patient agent to re-evaluate the situation and to select a new practice \textit{consulting an unknown doctor}.

The scenario manages the case in which the doctor directly introduces himself to the patient in the initial phase of the dialogue, as well as the case in which this greeting and presentation take place later.
Of course, the two alternatives produce different effects on the virtual patient. In the former case, a positive effect on the ``happiness" of the patient is achieved; instead in the latter case, an effect of ``displeasure" of the patient is obtained.
These two alternatives can be interpreted in the light of the social practices selected by the patient.
The practice contains a norm which states that the doctor and patient should introduce themselves if not known yet in order to show respect. If the doctor does not make this introduction at the start this gets a social meaning of showing disrespect to the patient. So the ``disappointment" of the patient could be seen as a consequence of the ``violation" of the social norm.
 
The Communicate! scenario envisages that the doctor provides more information to the patient about himself and his role within the hospital. In particular, the doctor is still in training with the patient's doctor. Knowing that the one who is in front of him is a doctor in training leads the patient agent to a state of ``anger'' if he is in an emotional state of ``contempt".
However, if the patient is in a state of ``happiness", he goes to a state of ``surprise".

The analysis of this portion of the dialogue shows that the effect of a communicative actions depends largely on the context in which they are uttered. Especially, the social meaning of the utterance depends on the social practice. Thus the ensuing emotional state of the patient also depends on this practice.

\section{Social practice agent architecture}

The structural dialogue state approach used in Communicate! is a viable solution to manage  scripted dialogues.
Although this is a good starting point, from an educational point of view it is important to allow the player to experience more varying and realistic dialogues.
In order to overcome the lack of flexibility of this approach, several approaches have been proposed in literature. 

For example, structural dialogue state approach has often been contrasted with plan-based approaches\cite{Cohen1997} \cite{DeMori1998}. Anyway, dialogue plan-based approaches require a great effort in the definition of commonsense knowledge and procedural processing, often leading to the creation of an opaque solution.

Moreover, to improve the educational effectiveness, the game should allow the user to play both the roles (doctor or patient), allowing the doctor to experience the patient point of view \cite{McCurdy}. To this aim, a design of the game that models both the agents roles is required.

In this work, the model of social practice is introduced in order to overcome the limitations of both structural dialogue state and plan-based approaches, sharing the motivation with the work of Traum \& Larsoon \cite{Traum2003}. 

\begin{figure*}[tb]
\begin{centering}
\includegraphics [scale=0.15]{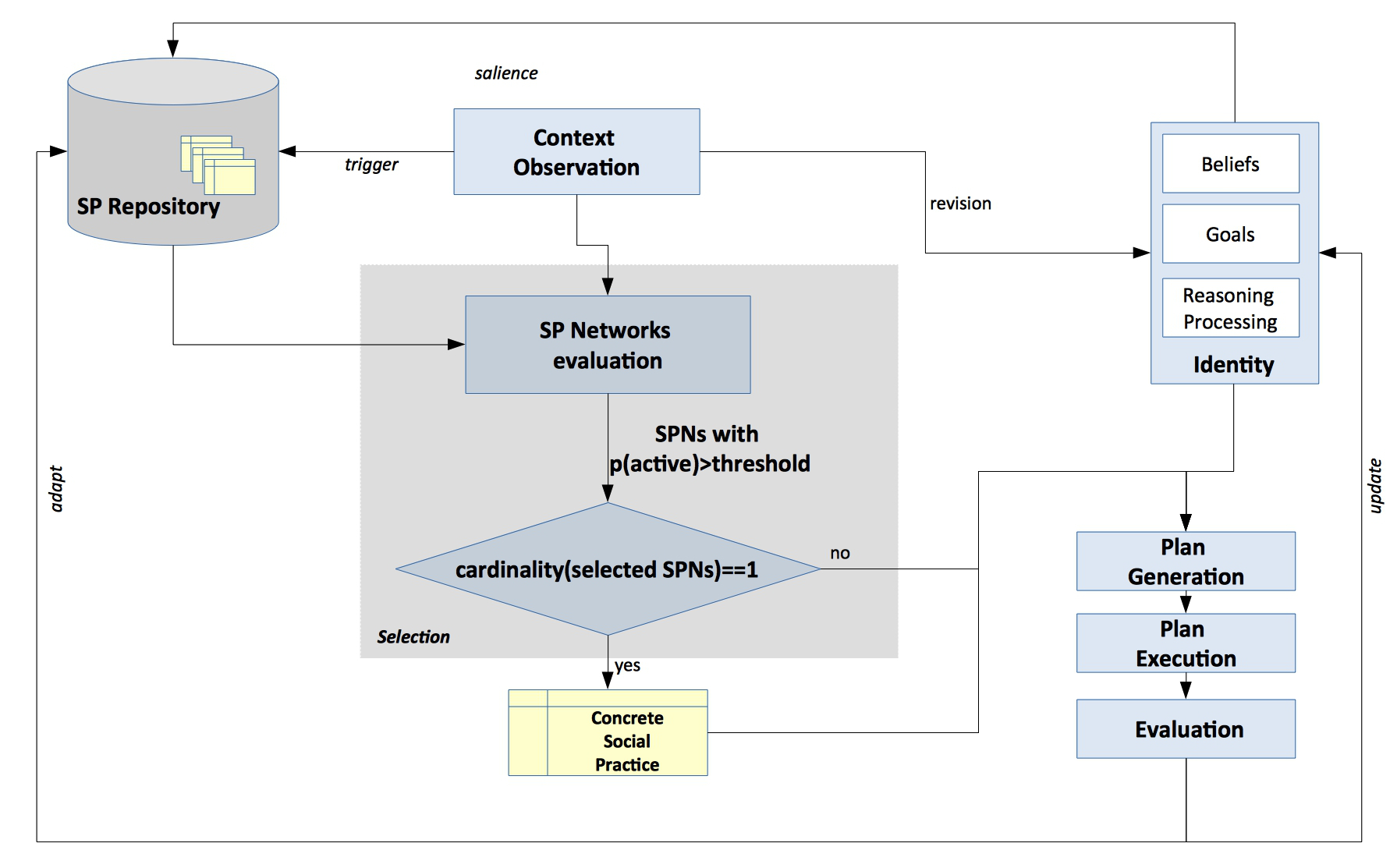}
\caption{Architecture for social reasoning  \cite{dignum}}
\label{fig:archi}
\end{centering}
\end{figure*}

The proposed agent's architecture (fig. \ref{fig:archi}) puts the identification of the social context at the basis of the deliberative process of the agents. The social context is modelled according to the theory of social practice (section \ref{sec:socialpractice}).
The identification of the most appropriate social practice allows for a dynamic management of the dialogue within the serious game, allowing the conversation to follow a specific but not rigid flow, and restricting considerably the number of possible sentences that may be issued by the agent and at the same time facilitating the understanding of the utterances of the interlocutor. 
Starting from the observation  of the context an agent updates his  beliefs and goals, and activates a set of possible social practices. He then triggers an internal reasoning process in order to select the most appropriate social practice (section \ref{sec:socialpracticeselection}).
Finally, according to the selected social practice he starts a deliberation process in order to manage the conversation within the context of that practice.
When the available information is not enough to activate a single social practice, the agent will interact with the user in order to obtain additonal information that discriminates the context further and leads to one social practice being the most suitable.

In this paper the focus is on the social practice modeling and selection, as highlighetd in the figure.

\subsection{The model of social practice}
\label{sec:socialpractice}

In this section the model of social practices described in \cite{dignum} is analysed in the context of a doctor-patient dialogue.
According to \cite{dignum}, the aim of the social practice model is to provide a representation scheme that allows the implementation of cognitive agents able to use the social practice as a first-class construct in the agent deliberation process.   In the following, the components of the social practice model are described by referring to a specific social practice (\textit{"Doctor Patient Dialogue"}) concerning the scenario of a doctor-patient interview. Table \ref{tab:social_practice} summarizes the components of the  social practice  under investigation.

\begin{table}[h!]
\centering
\caption{The doctor-patient dialogue as a social practice}
\label{tab:social_practice}
\begin{tabular}{|l|l|}

\hline

{\bf \parbox[t]{4cm}{Abstract Social Practice}}&{\bf Doctor Patient Dialogue}\\ \hline
Physical Context&\\ 
\multicolumn{1}{|r|}{Resources}&current time,medical instruments\\ 
\multicolumn{1}{|r|}{Places}& hospital,office\\ 
\multicolumn{1}{|r|}{Actors}& doctor1, patient1\\ \hline
Social Context&\\ 
\multicolumn{1}{|r|}{Social interpretation}&consulting room,consulting time,doctor has medical skills\\ 

\multicolumn{1}{|r|}{Roles}&doctor, patient, relative of the patient, nurse\\

\multicolumn{1}{|r|}{Norms}&
\parbox[t]{6cm}{patient is cooperative,\\ doctor is polite }\\
 \hline

Activities&\parbox[t]{8cm}{
welcome, presentation, patient's data gathering, patient symptom description \
constative(answer,confirm,disagree,agree)\\directive(ask,instruct,request)}\\ \hline

Plan patterns&
\multicolumn{1}{|c|}{
\includegraphics [scale=0.20]{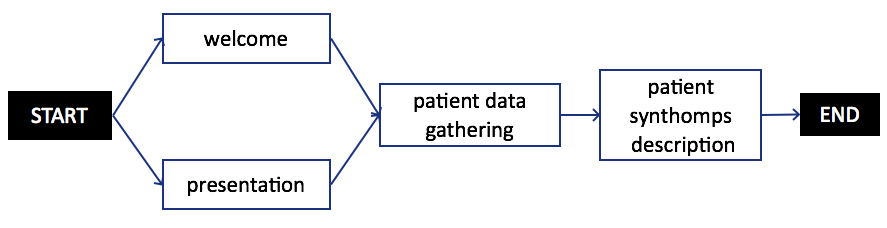}
}
\\ \hline
Meaning&
\parbox[t]{8cm}
{
support the patient, create trust, eliciting patient's problems and concerns, treatment and solution possible, empathic opportunity, empathic response
}

\\ \hline
Competences&
\parbox[t]{8cm}
{
listening effectively, being supportive and empathic, use effective explanatory
skills, adapt conversation to patient, discussing treatment options understandable
}
\\ \hline
\end{tabular}
\end{table}

\begin{itemize}
 
\item\textit{Physical Context} focuses on  the identification of the environment elements, that are both the physical  objects with a meaningful role in the practice (\textit{Resources}), the agents (human beings or autonomous systems) involved in the practice (\textit{Actors}) and the locations of objects and actors (\textit{Places}).
 
In the analyzed scenario  at least two people are expected to take part in the practice, the interview takes place in a hospital, inside a room  and at a specific time. Among the various resources available in this context, we can highlight the medical instruments visible in the room including the doctor's coat.

\item \textit{Social Context} describes the social interpretation of the elements sensed in the environment, in particular in \cite{dignum} a distinction is made between the so called \textit{Roles}, describing   specific behaviours that can be expected from specific actors, and  the interpretation of all the other elements, called \textit{Social Interpretation}. Moreover the \textit{Norms} identify  the rules that are expected inside a social practice. 
As an example, in the analyzed scenario, we can identify a physician role and a patient role,  and in some cases a relative of the patient.
The doctor-patient dialogue generally takes place during consultation hours in a special room used for the purpose. Finally, the type of clothing, and in particular the colour of the coat worn by the doctor are usually indicative of the "role" (doctor, assistant, nurse) that the professional has in the hospital. 
Different  norms can be related to the  roles of doctor and patient. As an example the doctor during the interview should totally reserve his attention to the patient; if suddenly there is an emergency,  the  norm is no more applicable. For what concerns  the role of the  patient it is important that he is cooperative.

\item \textit{Activities} are possible course of actions that agents can perform. In a fine-grained analysis, every communicative act of the agent can be considered as an activity. At this level we tie communicative acts to the speech act classification \cite{Searle} and not to specific utterances. Listing all the possible language productions is not possible and certainly not efficient; moreover, this allows us to organize the individual communicative acts in communication protocols.
Non communicative acts such as moves, gestures and so  on can be also considered at this level.
In a coarse-grain analysis, a set of simple actions can be considered as an activity; it is a sort of a scene in the plan of actions of the agent aimed to reach a specific sub-goal.
As an example a scene can be a set of communicative and non communicative  acts used  by the doctor to welcome the patient into the room.

\item  \textit{Plan Patterns}. A plan pattern is a pattern on the basis of which an agent can construct a plan to reach a goal. A plan pattern consists of an ordered set of scenes with a specific sub-goal that restrict the type of plans that can be used. 
As an example, the doctor will start and follow a specific protocol for the patient consultation. The protocol can be structured in the following scenes:  Welcome and Presentation,  Personal Data acquisition (biographic data, pathologies, medical treatments, allergies, ...),  Description of the symptoms.
The goal of the protocol is to start a doctor/patient relation and to proceed to the consult, however the situation can evolve and lead to quit (dotted lines in the image) the social practice. For example if the doctor acquiring the information recognizes that another doctor could be more competent for the patient's issues (in this case the doctor has not the competences required by the practice), or if the patient  shows an untrusted or disrespectful behavior (in this case a violation of a norm), the doctor can switch to another practice.

\item  \textit{Meaning}. This part of the social practice model, is used to define social meanings for the  agent's  activities and plans.
Considering communicative acts as activities, every act can assume a different meaning depending on the context of the dialogue, the cognitive status of the agents and so on.
As an example, communicative acts could be classified from an empathic point of view according to the classification proposed by \cite{suchman1997} (i.e. empathic opportunity, empathic response, etc...).
At the same time, the meaning of communicative acts includes its intended effects \cite{Clark}.
Of course these examples are not exhaustive of all the other different types of meaning that could be associated to the communicative acts.

\item  \textit{Competences} Competences are defined as the abilities an agent should have to perform the activities  of the social practice.
Amongst the medical skills required to interact with the patient there are several communication skills recommended to physicians \cite{maquire2002} \cite{Travaline2005}. For example, physicians should \textit{listen effectively}, \textit{being supportive and empathic}, \textit{use effective explanatory
skills}, \textit{discuss treatment options in a simple way}.
At the same time, patients should have health literacy skills in order to have an effective dialogue.

\end{itemize}

\subsection{A probabilistic Reasoner for Social Practice selection}
\label{sec:socialpracticeselection}

Starting from the observation  of the context and according to his identity the  agent activates a set of possible social practices. Based on his beliefs and goals it subsequently selects the most appropriate social practice. 
We propose a probabilistic reasoner for this selection process, because often not all information is available or certain. In those cases the agent will use its experience and derive as much information as possible to make a selection.

\begin{figure*}[tb]
\begin{centering}
\includegraphics [scale=0.30]{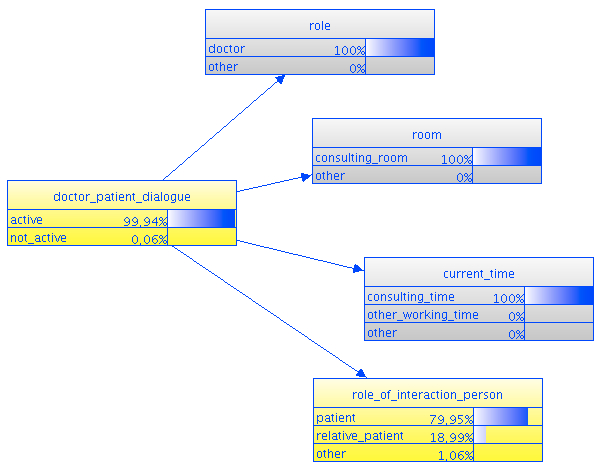}
\caption{The  decision network activation for the \textit{Doctor\_Patient\_Dialogue} social practice}
\vspace{-1.2em}
\label{fig:bn}
\end{centering}
\end{figure*}

The elements of the physical environment and their social interpretation related to a social practice, are represented in a causal graph. An example for the practice \textit{``Doctor Patient Dialogue"} is shown in figure \ref{fig:bn}. From the root node, representing the activation of the social practice, depart causal nodes representing its relevant elements (i.e. place, current time, ...). The states of these nodes are the possible social interpretations of such elements; for example the \textit{consulting time} is the social interpretation of the physical resource \textit{current time}. The causal links are expressed by means of conditional probability tables. 
The Bayesian networks allow the designer to adopt a top-down approach to formalize its initial knowledge about the social practice. In later versions  learning algorithms will be implemented to allow the agent to learn the  tables of conditional probabilities related to the social practices, but we leave that out for the moment. 
The algorithm of social practice selection is shown in  figure \ref{fig:archi}, and explained below according to the  scenario described in section \ref{sec:training}.
First, the observation of the context and the identity of the agent produces an initial selection of practices. If the agent has the role of doctor and he is at the hospital,  possible social practices are for example those related to consulting patients  and the management of an emergency. According to the available information it is possible to set evidences in the causal graphs associated with the different social practices. 
In particular if the \textit{current\_time} is a \textit{consulting\_time} and the doctor is inside a consulting\_room, the analysis of the  activation probabilities of the possible practices shows that the practice related to a patient consultation 
(\textit{doctor\_ patient\_dialogue}) has the highest probability, as shown in figure \ref{fig:bn}. In particular even if the agent does not yet know who is his interlucotor, according to the model of the activated social practice, he estimates that he should be a patient with a high probability.

When the situation evolves several things might happen. The person either can be indeed a patient, an hospital employee (is wearing some hospital stuff), or can be not alone.
Depending on the actual situation the expectations of the social practice are confirmed and the practice is continued or the social practice is re-evaluated with the new information. 
When a social practice is chosen 
the agent can start a further context analysis in order to discriminate among more concrete social practices that lead to more specific plan patterns, roles, norms etc.

\section{Conclusion}
In this paper we have given some initial steps showing the usability of social practices for managing dialogues in serious games. Mainly we have shown that social practice captures both the physical as well as the social context and thus forms an important source of information for the interpretation and generation of speech acts. Having this context available as source of information allows for a more accurate interpretation of the utterances as they can be compared to the expected type of utterances according to the practice.
Because social practices do not enforce a fixed script but rather indicate usable plan patterns they allow for flexibility in the dialogue while still ensuring a (joint) goal to be achieved.

Of course, there are many open questions such as the connections between different social practices, the learning and adaptation of the agent and we have also not shown the actual dialogue manager that is used by the agent to generate the actual utterances.
The first steps are to fully implement the agent architecture and using a simple dialogue planner as in Communicate! produce at least the same dialogues but on the fly. Moreover, the Information State approach \cite{Traum2013} will be analyzed in order to verify if a dialogue manager defined according to this approach could be used as a valuable deliberation engine for an agent compliant with the social practice model.

\section{Acknowledgements}
\label{sec:Acknowledgements}
We are grateful to Prof. J. Jeuring , for his useful explanation of the Communicate! project and for providing us the scenarios of the game.

%
%

\end{document}